\documentclass{article}

\usepackage[final,nonatbib]{assets/techreport}

\usepackage[utf8]{inputenc} 
\usepackage[T1]{fontenc}    
\usepackage{hyperref}       
\usepackage{url}            
\usepackage{booktabs}       
\usepackage{amsfonts}       
\usepackage{nicefrac}       
\usepackage{microtype}      
\usepackage{xcolor}         
\usepackage{pgfplots}
\usepackage{subcaption}
\usepackage{csvsimple}
\usepackage{siunitx}
\usepackage[flushleft]{threeparttable} 
\usepackage{pgfplotstable}
\usepackage{enumitem}

\title{VitalLens: Take A Vital Selfie}

\author{%
  Philipp V.~Rouast \\
  Rouast Labs\\
  \texttt{philipp@rouast.com} \\
}

\csvreader[before reading=\def\totalparticipantsprosit{0}\def\totalchunksprosit{0}\def\totaltimeprosit{0}\def\totalsessionsprosit{0}]{data/prosit_summary.csv}{participants=\participants,chunks=\chunks,time=\time,sessions=\sessions}{%
	\pgfmathsetmacro{\totalparticipantsprosit}{\totalparticipantsprosit+\participants}%
	\pgfmathsetmacro{\totalchunksprosit}{\totalchunksprosit+\chunks}%
	\pgfmathsetmacro{\totaltimeprosit}{\totaltimeprosit+\time}%
	\pgfmathsetmacro{\totalsessionsprosit}{\totalsessionsprosit+\sessions}%
}

\csvreader[before reading=\def\totalparticipantsvvmedium{0}\def\totalchunksvvmedium{0}\def\totaltimevvmedium{0}]{data/vv_medium_summary.csv}{participants=\participants,chunks=\chunks,time=\time}{%
	\pgfmathsetmacro{\totalparticipantsvvmedium}{\totalparticipantsvvmedium+\participants}%
	\pgfmathsetmacro{\totalchunksvvmedium}{\totalchunksvvmedium+\chunks}%
	\pgfmathsetmacro{\totaltimevvmedium}{\totaltimevvmedium+\time}%
}

\csvreader[before reading=\def\totalparticipantsvvafrica{0}\def\totalchunksvvafrica{0}\def\totaltimevvafrica{0}]{data/vv_africa_small_summary.csv}{participants=\participants,chunks=\chunks,time=\time}{%
	\pgfmathsetmacro{\totalparticipantsvvafrica}{\totalparticipantsvvafrica+\participants}%
	\pgfmathsetmacro{\totalchunksvvafrica}{\totalchunksvvafrica+\chunks}%
	\pgfmathsetmacro{\totaltimevvafrica}{\totaltimevvafrica+\time}%
}

\begin{document}
\begin{filecontents*}{data/prosit_summary.csv}
split,participants,sessions,chunks,time
Train,114,124,6765,19.4
Valid,23,26,1599,4.5
Test,20,23,1403,3.9
\end{filecontents*}

\begin{filecontents*}{data/vv_medium_summary.csv}
split,participants,chunks,time
Test,289,1108,4.6
\end{filecontents*}

\begin{filecontents*}{data/vv_africa_small_summary.csv}
split,participants,chunks,time
Train,79,158,0.5
Valid,25,50,0.1
Test,25,50,0.1
\end{filecontents*}

\maketitle

\begin{abstract}
This report introduces VitalLens, an app that estimates vital signs such as heart rate and respiration rate from selfie video in real time.
VitalLens uses a computer vision model trained on a diverse dataset of video and physiological sensor data.
We benchmark performance on several diverse datasets, including VV-Medium, which consists of \sisetup{round-mode=places,round-precision=0}\num{\totalparticipantsvvmedium} unique participants.
VitalLens outperforms several existing methods including POS and MTTS-CAN on all datasets while maintaining a fast inference speed.
On VV-Medium, VitalLens achieves mean absolute errors of 0.71 bpm for heart rate estimation, and 0.76 bpm for respiratory rate estimation.
\end{abstract}

\section{Introduction}
\label{sec:introduction}

Video of the human face and upper body has proven to be a rich source of information about an individual's physiological state \cite{verkruysse2008remote}.
In particular, signals embedded in these videos can be harnessed to estimate vital signs through a process known as Remote Photoplethysmography (rPPG).
This capability holds immense potential for non-invasive and real-time health monitoring applications.

Various rPPG approaches have been proposed, including handcrafted algorithms and ones learned from empirical data.
Handcrafted algorithms offer advantages of fast inference speed and no training data required, but usually lack in accuracy; notable approaches include the original approach G \cite{verkruysse2008remote}, and the more sophisticated CHROM \cite{de2013robust} and POS \cite{wang2017algorithmic}.
Given enough high-quality data to learn from, learning-based approaches such as DeepPhys \cite{chen2018deep} and MTTS-CAN \cite{liu2020multi} hold the promise of greater accuracy, which is traded off against inference speed.
There are also other learning-based approaches, which we do not cover here as they are not aiming to achieve real-time inference.

In this context, we introduce VitalLens, an innovative application designed to estimate vital signs, including heart rate and respiratory rate, in real time from selfie videos.
This report focuses on the capabilities and limitation of VitalLens.
Our key contributions are:

\begin{itemize}
	\item \textbf{Real-time rPPG application.} We introduce VitalLens -- the first widely distributed rPPG application for real time estimation of vitals such as of heart rate and respiratory rate.\footnote{\href{https://apps.apple.com/us/app/vitallens/id6472757649}{Click to get VitalLens for iOS.}} 
	\item \textbf{Comprehensive evaluation on a large dataset.} We evaluate the performance of VitalLens on multiple datasets, including VV-Medium \cite{toye2023vital} with \sisetup{round-mode=places,round-precision=0}\num{\totalparticipantsvvmedium} participants. VitalLens outperforms G, CHROM, POS, DeepPhys, and MTTS-CAN in terms of estimation accuracy. On VV-Medium, VitalLens achieves mean absolute errors of 0.71 bpm for heart rate estimation, and 0.76 bpm for respiratory rate estimation, while taking 18 ms to run inference.
	\item \textbf{Analysis of factors impacting performance.} We systematically investigate factors impacting estimation performance. We find that \textit{participant movement} and \textit{variation of participant illuminance} are the main factors impacting performance. Compared to approaches like POS, VitalLens is less impacted by darker skin types when estimating heart rates. 
	\item \textbf{Maintaining end-user privacy}. VitalLens runs inference locally, meaning that video and vitals never leave the device.
\end{itemize}

\section{Architecture}
\label{sec:architecture}

\begin{figure}[h!]
	\centering
 	\includegraphics[width=\textwidth]{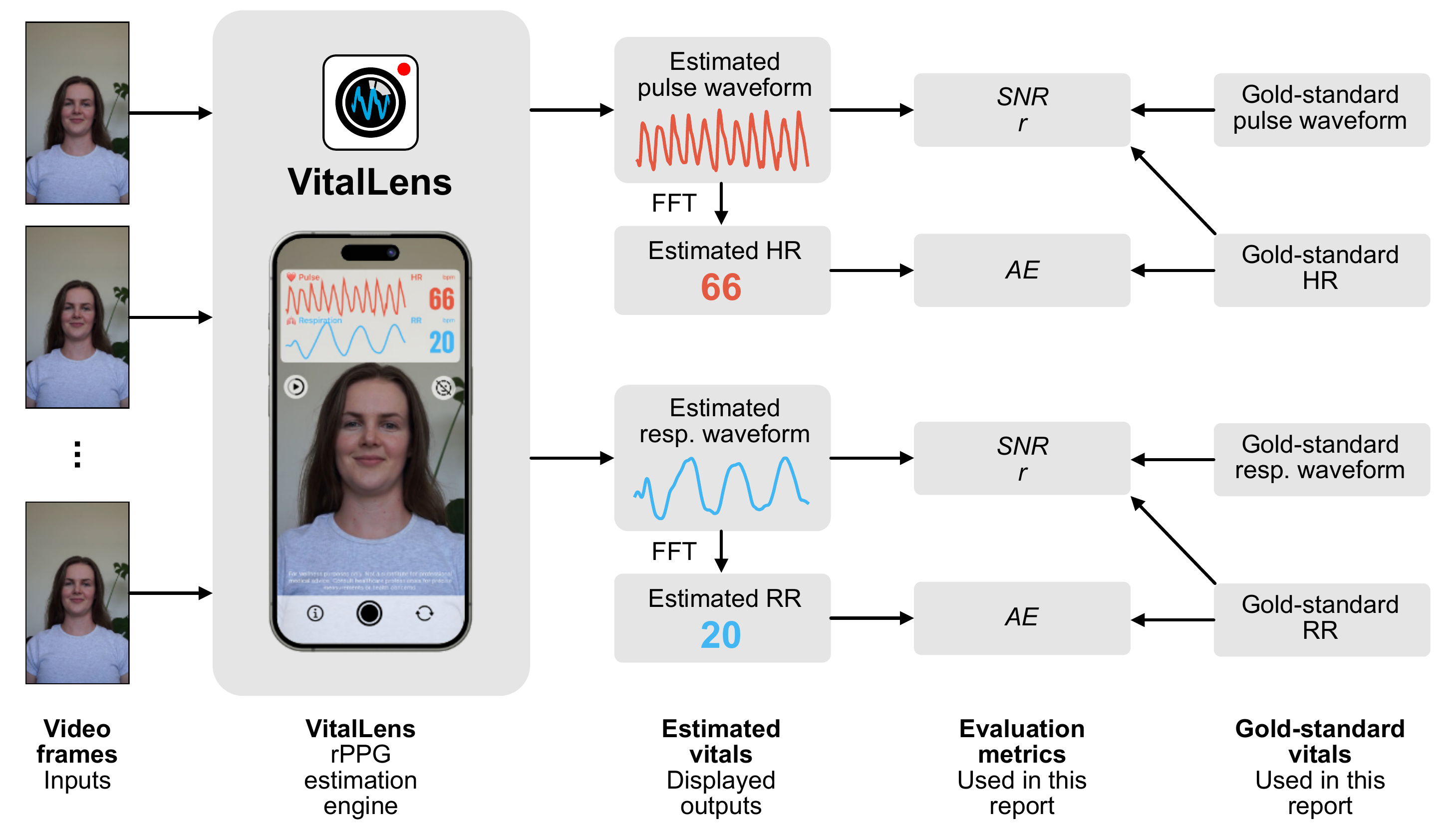}
 	\caption{VitalLens estimates vital sign waveforms from video frames. The app displays the video feed with an overlay of the estimated pulse and respiration waveforms as well as the derived heart rate and respiration rate. We report evaluation metrics for all displayed quantities by comparing the estimations with gold-standard labels.}
 	\label{fig:overview}
\end{figure}

VitalLens takes a series of video frames as inputs, and uses them to estimate the pulse waveform and the respiration waveform.
The corresponding heart rate (HR) and respiratory rate (RR) estimates are derived from these waveforms using Fast Fourier Transform (FFT).
Please refer to Figure \ref{fig:overview} for a broad overview of how VitalLens operates, along with the evaluation method used in Section \ref{sec:results}.

The estimation engine built into VitalLens (henceforth ``VitalLens'') is a computer vision model trained on a diverse dataset of video and gold-standard ground truth physiological data.
The model architecture is broadly based on the \textit{EfficientNetV2} \cite{tan2021efficient} model family, enhanced with several improvements in architecture and model optimization to enable efficient training and inference in the rPPG domain.
This report does not contain any further detail about the architecture of the model or training methodology - we reserve this for future publication.

\section{Datasets}
\label{sec:datasets}

VitalLens is trained on the \textit{PROSIT} and \textit{Vital Videos Africa} datasets.
The vast majority of training data comes from PROSIT.
For evaluation, we use the entire \textit{Vital Videos Medium} dataset as well as the test sets of the \textit{PROSIT} and \textit{Vital Videos Africa} datasets.

\subsection{PROSIT}

PROSIT (\underline{P}hysiological \underline{R}ecordings \underline{O}f \underline{S}ubjects using \underline{I}maging \underline{T}echnology) is our in-house dataset collected in Australia for practical rPPG applications.

\paragraph{Participant recruitment and session protocol.} 
As part of our goal of creating a diverse rPPG dataset for practical applications, we recruit participants and collect data at various locations such as residential homes, offices, libraries, and clubs.
Each potential participant was required to go through an informed consent process in accordance with Australian privacy laws prior to participation.
During the session, participants are asked to complete tasks on a hand-held iPad while sensor data is being recorded.
Participants are not explicitly asked to remain stationary, which results in varying amounts of participant movement.
For some sessions, the camera is on a tripod and thereby stationary, while for other sessions the camera is fixed to the iPad and thereby not necessarily stationary.

\paragraph{Sensors and collected data.}
The time-synchronized sensor array used for PROSIT consists of a video camera, electrocardiogram (ECG), pulse oximetry, blood pressure monitor, and an ambient light sensor.
This yields a rich set of data including video, ECG, PPG, SpO2, respiration, blood pressure, and ambient luminance.
We also collect age, gender, height, and skin type metadata according to the Fitzpatrick scale \cite{fitzpatrick1975soleil}.

\paragraph{Pre-processing.}
We pre-process and split each session into small chunks of 5-20 seconds with valid video and signals.
As part of this step, we also calculate summary vitals for each chunk from the continuous signals, and extract further metadata such as the amount of participant movement and illuminance variation.

\paragraph{Dataset size and split.}
Development of PROSIT is ongoing.
As of the writing of this report, it comprises \sisetup{round-mode=places,round-precision=0}\num{\totalparticipantsprosit} unique participants across \sisetup{round-mode=places,round-precision=0}\num{\totalsessionsprosit} recording sessions in 45 different locations.
This results in a total of \sisetup{round-mode=places,round-precision=0}\num{\totalchunksprosit} chunks or \sisetup{round-mode=places,round-precision=1}\num{\totaltimeprosit} hours of data.

\begin{table}[h!]
 	\caption{PROSIT Dataset Size}
 	\label{tab:prosit-summary}
 	\centering
  	\csvreader[
		tabular=llll,
    	table head=\toprule Split & \# Participants & \# Chunks & Time \\ \midrule,
    	table foot=\midrule Total & \sisetup{round-mode=places,round-precision=0}\num{\totalparticipantsprosit} & \sisetup{round-mode=places,round-precision=0}\num{\totalchunksprosit} & \sisetup{round-mode=places,round-precision=1}\SI{\totaltimeprosit}{\hour} \\ \bottomrule,
    	late after line=\\,
    	before reading={\catcode`\%=12}, after reading={\catcode`\%=14}
  	]{data/prosit_summary.csv}{}%
  {\csvcoli & \num{\csvcolii} & \num{\csvcoliv} & \SI{\csvcolv}{\hour} }
\end{table}

Each participant is randomly assigned to be part of either the \textit{training}, \textit{validation}, or \textit{test} set to ensure that all participants seen during validation and test are previously unseen by the model.

\subsection{Vital Videos Medium}

Vital Videos is a large, diverse dataset for rPPG collected in Belgium \cite{toye2023vital}.
It is the largest dataset we are aware of that is available for research without academic affiliation.
We use a slightly extended version of the medium instantiation (``VV-Medium''), which consists of \sisetup{round-mode=places,round-precision=0}\num{\totalparticipantsvvmedium} participants.
Both camera and participants are stationary in this dataset.

\paragraph{Pre-processing.}

We pre-process VV-Medium using the same steps applied to PROSIT.
As part of this step, we create small chunks, calculate summary vitals, and extract further metadata.
Note that for most chunks, the missing respiratory signal was synthetically created.\footnote{This was done using an earlier version of our model trained on PROSIT. We then manually verified the correctness by visual inspection of both the label and video.}

\begin{table}[h!]
 	\caption{VV-Medium Dataset Size}
 	\label{tab:vv-medium-summary}
 	\centering
  	\csvreader[
		tabular=llll,
    	table head=\toprule Split & \# Participants & \# Chunks & Time \\ \midrule,
    	table foot=\bottomrule,
    	late after line=\\,
    	before reading={\catcode`\%=12}, after reading={\catcode`\%=14}
  	]{data/vv_medium_summary.csv}{}%
  {\csvcoli & \num{\csvcolii} & \num{\csvcoliii} & \sisetup{round-mode=places,round-precision=1}\SI{\csvcoliv}{\hour} }
\end{table}

The entirety of VV-Medium is used to test the capabilities of VitalLens.

\subsection{Vital Videos Africa}

Vital Videos Africa is a new dataset from the authors of Vital Videos aiming to address the insufficient share of participants with skin types 5 and 6 usually found in rPPG datasets.
It was collected in Ghana, with the majority of participants having skin type 5 or 6.
We use a small instantiation (``VV-Africa-Small''), which consists of \sisetup{round-mode=places,round-precision=0}\num{\totalparticipantsvvafrica} participants.
Both camera and participants are stationary in this subset.

\paragraph{Pre-processing.}

We pre-process VV-Africa-Small using the same steps applied to PROSIT.
As part of this step, we create small chunks, calculate summary vitals, and extract further metadata.
Note that for some chunks, the missing respiratory signal was synthetically created using the same procedure as for VV-Medium.

\begin{table}[h!]
 	\caption{VV-Africa-Small Dataset Size}
 	\label{tab:vv-africa-small-summary}
 	\centering
  	\csvreader[
		tabular=llll,
    	table head=\toprule Split & \# Participants & \# Chunks & Time \\ \midrule,
    	table foot=\midrule Total & \sisetup{round-mode=places,round-precision=0}\num{\totalparticipantsvvafrica} & \sisetup{round-mode=places,round-precision=0}\num{\totalchunksvvafrica} & \sisetup{round-mode=places,round-precision=1}\SI{\totaltimevvafrica}{\hour} \\ \bottomrule,
    	late after line=\\,
    	before reading={\catcode`\%=12}, after reading={\catcode`\%=14}
  	]{data/vv_africa_small_summary.csv}{}%
  {\csvcoli & \num{\csvcolii} & \num{\csvcoliii} & \sisetup{round-mode=places,round-precision=1}\SI{\csvcoliv}{\hour} }
\end{table}

Each participant is randomly assigned to be part of either the \textit{training}, \textit{validation}, or \textit{test} set to ensure that all participants seen during validation and test are previously unseen by the model.

\subsection{Final training dataset: PROSIT + VV-Africa-Small}
\begin{filecontents*}{data/training_summary.csv}
source,participants,chunks,time
PROSIT,114,6765,19.4
VV-Africa-Small,79,158,0.5
\end{filecontents*}
\csvreader[before reading=\def\totalparticipantstraining{0}\def\totalchunkstraining{0}\def\totaltimetraining{0}]{data/training_summary.csv}{participants=\participants,chunks=\chunks,time=\time}{%
		\pgfmathsetmacro{\totalparticipantstraining}{\totalparticipantstraining+\participants}%
		\pgfmathsetmacro{\totalchunkstraining}{\totalchunkstraining+\chunks}%
		\pgfmathsetmacro{\totaltimetraining}{\totaltimetraining+\time}%
	}
We combine the training sets of PROSIT and VV-Africa-Small for training.
As of this writing, this makes a total of \sisetup{round-mode=places,round-precision=0}\num{\totalparticipantstraining} unique participants, \sisetup{round-mode=places,round-precision=0}\num{\totalchunkstraining} chunks or \sisetup{round-mode=places,round-precision=1}\num{\totaltimetraining} hours of data.

\begin{table}[h!]
 	\caption{VitalLens Training Dataset Size}
 	\label{tab:training-summary}
 	\centering
  	\csvreader[
		tabular=llll,
    	table head=\toprule Source & \# Participants & \# Chunks & Time \\ \midrule,
    	table foot=\midrule Total & \sisetup{round-mode=places,round-precision=0}\num{\totalparticipantstraining} & \sisetup{round-mode=places,round-precision=0}\num{\totalchunkstraining} & \sisetup{round-mode=places,round-precision=1}\SI{\totaltimetraining}{\hour} \\ \bottomrule,
    	late after line=\\,
    	before reading={\catcode`\%=12}, after reading={\catcode`\%=14}
  	]{data/training_summary.csv}{}%
  {\csvcoli & \num{\csvcolii} & \num{\csvcoliii} & \sisetup{round-mode=places,round-precision=1}\SI{\csvcoliv}{\hour} }
\end{table}

\paragraph{Dataset demographics.}
The demographics of our training dataset are given in Figure \ref{fig:training-demographics}.
The participants are predominantly on the younger side, but as we show in Section \ref{sec:results}, this is not an issue in practice.
Genders are equally represented.
Although the skin type diversity of PROSIT by itself is lacking, this combined training dataset has a diverse representation of skin types.
As we show in Section \ref{sec:results}, this helps us to reduce the skin type bias of VitalLens.

\definecolorseries{myseries1}{rgb}{step}[rgb]{1,0,0}{-1,0,1}
\resetcolorseries{myseries1}%

\definecolorseries{myseries2}{rgb}{step}[rgb]{1.0, 0.75, 0.65}{-0.134,-0.122,-0.13}
\resetcolorseries{myseries2}%

\newcommand{\slice}[5]{
	\pgfmathsetmacro{\midangle}{0.5*#1+0.5*#2}
	\begin{scope}
		\clip (0,0) -- (#1:1) arc (#1:#2:1) -- cycle;
		\colorlet{SliceColor}{#3!!+}
		\fill[inner color=SliceColor!30,outer color=SliceColor!60] (0,0) circle (1cm);
	\end{scope}
	\draw[thick] (0,0) -- (#1:1) arc (#1:#2:1) -- cycle;
	\node[label={[font=\small]\midangle:#5}] at (\midangle:1) {};
	\pgfmathsetmacro{\temp}{min((#2-#1-10)/110*(-0.3),0)}
	\pgfmathsetmacro{\innerpos}{max(\temp,-0.5) + 0.8}
	\node[font=\small] at (\midangle:\innerpos) {#4};
}

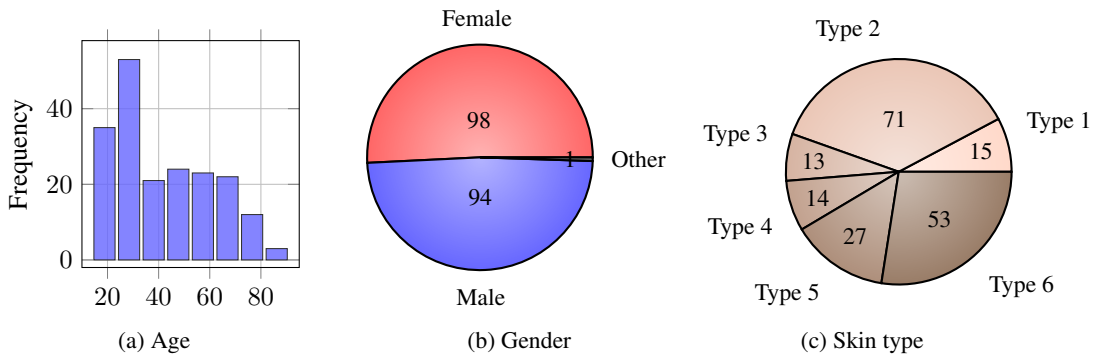
\begin{figure}[h!]
	\centering
    \begin{subfigure}{0.32\textwidth}
    	\centering
    	\begin{tikzpicture}
        	\begin{axis}[
            		table/col sep=comma,
            		ybar,
            		xmin=10,
            		xmax=95,
            		ylabel=Frequency,
            		width=\textwidth,
            		height=4.6cm,
            		grid=major,
            		bar width=8pt,
            		y label style={yshift=-12pt},
            		label style={inner sep=0pt}
            	]
            	\addplot[fill=blue!60,opacity=0.8] table[x index=0, y index=1] {data/age_histogram.csv};
        	\end{axis}
    	\end{tikzpicture}
    	\caption{Age}
    \end{subfigure}
    \hfill
    \begin{subfigure}{0.32\textwidth}
    	\centering
    	\begin{tikzpicture}[scale=1.5]%
    	\def\mya{0}\def\myb{0}
    	\begin{filecontents*}{data/gender.csv}
Label,Value
Female,98
Male,94
Other,1
\end{filecontents*}
\csvreader[before reading=\def\mysum{0}]{data/gender.csv}{Value=\Value}{%
			\pgfmathsetmacro{\mysum}{\mysum+\Value}%
		}
		\csvreader[head to column names]{data/gender.csv}{}{%
			\let\mya\myb
			\pgfmathsetmacro{\myb}{\myb+\Value}
			\slice{\mya/\mysum*360}{\myb/\mysum*360}{myseries1}{\Value}{\Label}
		}
		\end{tikzpicture}%
		\caption{Gender}
    \end{subfigure}
    \begin{subfigure}{0.32\textwidth}
    	\centering
    	\begin{tikzpicture}[scale=1.5]%
		\begin{filecontents*}{data/skin_type.csv}
Label,Value
Type 1,15
Type 2,71
Type 3,13
Type 4,14
Type 5,27
Type 6,53
\end{filecontents*}
\csvreader[before reading=\def\mysum{0}]{data/skin_type.csv}{Value=\Value}{%
			\pgfmathsetmacro{\mysum}{\mysum+\Value}%
		}
    	\def\mya{0}\def\myb{0}
		\csvreader[head to column names]{data/skin_type.csv}{}{%
			\let\mya\myb
			\pgfmathsetmacro{\myb}{\myb+\Value}
			\slice{\mya/\mysum*360}{\myb/\mysum*360}{myseries2}{\Value}{\Label}
		}
		\end{tikzpicture}%
		\caption{Skin type}
    \end{subfigure}
    \hfill
    \caption{Participant demographics in training dataset}
    \label{fig:training-demographics}
\end{figure}

\begin{figure}[h!]
    \centering
    \begin{subfigure}{0.48\textwidth}
    	\centering
    	\begin{tikzpicture}
        	\begin{axis}[
            		table/col sep=comma,
            		ybar,
            		ylabel=Frequency,
            		xlabel={Heart rate [bpm]},
            		xmin=35,
            		xmax=130,
            		width=\textwidth,
            		height=5cm,
            		grid=major,
            		bar width=12pt
            	]
            	\addplot[fill=red!60,opacity=0.8] table[x index=0, y index=1] {data/hr_histogram.csv};
        	\end{axis}
    	\end{tikzpicture}
    \end{subfigure}
    \hfill
    \begin{subfigure}{0.48\textwidth}
    	\centering
    	\begin{tikzpicture}
        	\begin{axis}[
            		table/col sep=comma,
            		ybar,
            		ylabel=Frequency,
            		xlabel={Respiratory rate [bpm]},
            		xmin=-2,
            		xmax=40,
            		width=\textwidth,
            		height=5cm,
            		grid=major,
            		bar width=12pt
            	]
            	\addplot[fill=blue!60,opacity=0.8] table[x index=0, y index=1] {data/rr_histogram.csv};
        	\end{axis}
    	\end{tikzpicture}
    \end{subfigure}
    \caption{Distributions of chunk summary vitals in training dataset}
    \label{fig:training-vitals-histogram}
\end{figure}
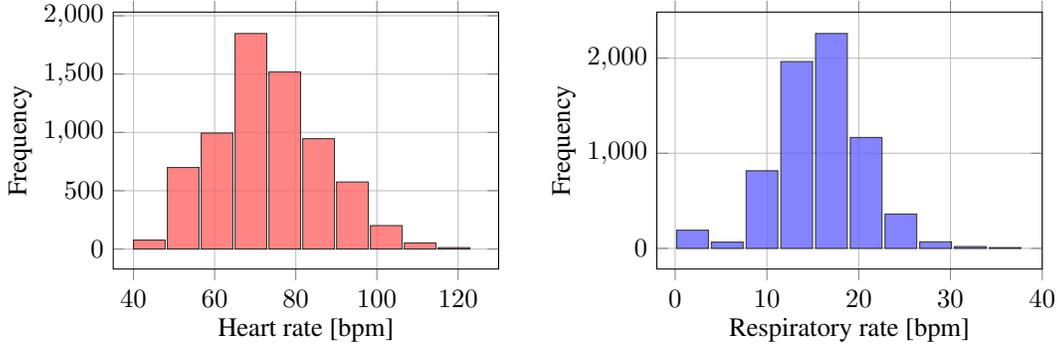

\paragraph{Dataset vitals diversity.}
Distributions of the vitals in our training dataset are given in Figure \ref{fig:training-vitals-histogram}.
The participants are mostly healthy, so these vitals fall in the typical ranges.
There are several participants who have an irregular heartbeat.

\section{Methodology}
\label{sec:methodology}

To develop and evaluate VitalLens, we follow a systematic approach involving model training, validation, and testing:

\paragraph{Model Training.}
The training process optimizes the model parameters to learn the mapping between facial video signals and ground truth physiological data.
For this purpose, we use a diverse training dataset of \sisetup{round-mode=places,round-precision=0}\num{\totalparticipantstraining} unique participants introduced in Section \ref{sec:datasets}.

\paragraph{Model Validation.}
While developing the model, we train many different versions of our model.
During training, we continuously monitor the model's performance on the validation sets of PROSIT and VV-Africa-Small as outlined in Section \ref{sec:datasets}.
Since the training dataset is disjoint from both validation sets in terms of participants, this allows us to calculate validation metrics which measure how well each model has learned to generalize in estimating vital signs from video.
We used the validation metrics to select the final model to be used in VitalLens.

\paragraph{Model Testing.}
The final evaluation of VitalLens is conducted by benchmarking the generalization abilities of the final model on several datasets, all of which are participant-disjoint from both the training and validation datasets:

\begin{itemize}
	\item The entire VV-Medium dataset, a large and diverse dataset of \sisetup{round-mode=places,round-precision=0}\num{\totalparticipantsvvmedium} participants introduced in Section \ref{sec:datasets}. It includes participants with varying demographics and environmental conditions.
	\item The test set of the PROSIT dataset. This is important as PROSIT includes camera and participant movement, which is to be expected when VitalLens is used in the wild.
	\item The test set of the VV-Africa-Small dataset.
\end{itemize}

\paragraph{Inference and Privacy.}
When end-users use VitalLens on their devices, inference runs locally.
This means that the video and vitals of end-users never leave their devices, which is important to maintain privacy.
Another benefit is that an internet connection is not required to run VitalLens.

\section{Results and Discussion}
\label{sec:results}

We evaluated VitalLens on three datasets - VV-Medium, the PROSIT test set, and the VV-Africa-Small test set.
None of the participants included in these datasets were previously seen by VitalLens.
In addition, we also evaluated several other existing methods to allow benchmarking of VitalLens:
The handcrafted methods G, CHROM, and POS, as well as the learning-based methods DeepPhys and MTTS-CAN.
To allow a fair comparison, we trained the latter two on the same training data as VitalLens.

Furthermore, we conduct a regression analysis by using metadata to predict the performance of VitalLens.
This helps us identify the factors affecting performance, and informs ``how to use'' advice offered to users of VitalLens.

Finally, we analyze the influence of the factors of interest in more detail.

\subsection{Vitals estimation}

Vitals estimation is performed separately for each chunk at 30 frames per second for all compared methods.
As shown in Figure \ref{fig:overview}, we use the dataset labels and model estimations to calculate several evaluation metrics for each chunk:

\begin{itemize}
	\item \textbf{Absolute error} (AE, smaller is better): Used to measure the absolute difference between the estimated rate and the true rate.
	\item \textbf{Signal-to-noise ratio} (SNR, larger is better): Measures the ratio of the true rate signal to the noise present in the estimated waveform in decibels.
	\item \textbf{Pearson correlation coefficient} ($r$, larger is better): Measures the the linear correlation between the estimated waveform and the true waveform.
\end{itemize}

We then report the mean for each of these metrics across all chunks in the given dataset.
As is common, we denote the mean AE as MAE as is common, but continue to use the abbreviations SNR and $r$ when reporting the mean SNR and mean $r$.

\newcommand{\printMetric}[1]{%
  \ifthenelse{\equal{#1}{0.0}}%
    {--}%
    {\sisetup{round-mode=places,round-precision=2}\num{#1}}%
}

\begin{table}[h!]
 	\caption{Vitals estimation results on VV-Medium}
 	\label{tab:results-vv-medium}
 	\centering
 	\begin{threeparttable}	
  	\begin{filecontents*}{data/results_vv_medium.csv}
method,hr_mae,pulse_snr,pulse_cor,rr_mae,resp_snr,resp_cor,inf_speed
G,10.087834803497907,-3.5232535025890064,0.08314889718528289,0.0,0.0,0.0,3.4
CHROM,5.184493803606818,0.48478767952133506,0.32671850649955436,0.0,0.0,0.0,4.2
POS,2.2443485127432874,5.973038016653952,0.1007668282232118,0.0,0.0,0.0,3.6
DeepPhys,9.624271143824076,-2.0668813013961147,0.5223716541855935,2.0636525535012673,4.215953058238846,0.7252385136537911,9.8
MTTS-CAN,1.310090833365592,6.29519415454263,0.7127485536377168,5.3616697116947725,-4.25451418906016,0.43093354471018813,22.1
VitalLens,0.7077107974109269,11.018533710859652,0.7456786230652699,0.7624069768817882,7.851042020004585,0.8924095442410327,18.0
\end{filecontents*}
\csvreader[
		tabular=lccccccc,
    	table head=\toprule Method & \multicolumn{1}{c}{Heart rate} & \multicolumn{2}{c}{Pulse wave} & \multicolumn{1}{c}{Respiratory rate\tnote{a}} & \multicolumn{2}{c}{Respiration wave\tnote{a}} & Inference \\ & MAE $\downarrow$ & SNR $\uparrow$ & $r \uparrow$ & MAE $\downarrow$ & SNR $\uparrow$ & $r \uparrow$ & time in ms\tnote{b} \\ \midrule,
    	table foot=\bottomrule,
    	late after line=\\,
    	before reading={\catcode`\%=12}, after reading={\catcode`\%=14}
  	]{data/results_vv_medium.csv}{}%
  {\csvcoli & \printMetric{\csvcolii} & \printMetric{\csvcoliii} & \printMetric{\csvcoliv} & \printMetric{\csvcolv} & \printMetric{\csvcolvi} & \printMetric{\csvcolvii} & \sisetup{round-mode=places,round-precision=1}\num{\csvcolviii}}
  \begin{tablenotes}
	\item[a] Missing respiratory signals were synthetically created and manually verified.
	\item[b] Time to run inference for a single video frame.
	\end{tablenotes}
  \end{threeparttable}
\end{table}

Results for VV-Medium are reported in Table \ref{tab:results-vv-medium}.
The heart rate and respiratory rate estimated by VitalLens across all chunks was off by 0.71 bpm and 0.76 bpm on average from the gold-standard labels, while taking 18 ms for inference.

\begin{figure}[h!]
  \centering
  \begin{subfigure}{0.48\textwidth}
    \begin{tikzpicture}
      \begin{axis}[
          xlabel=Gold-standard heart rate,
          ylabel=Estimated heart rate,
          height=7cm,
          width=\textwidth,
      ]
      \addplot[
          only marks,
          mark=o,
          color=red,
      ] table [col sep=comma, x=hr_true, y=hr_est] {data/results_vv_medium_scatter.csv};
      \addplot [gray,line width=0.2, domain=35:130, samples=2] {x};
      \end{axis}
    \end{tikzpicture}
  \end{subfigure}
  \hfill
  \begin{subfigure}{0.48\textwidth}
    \begin{tikzpicture}
      \begin{axis}[
          xlabel=Gold-standard respiratory rate,
          ylabel=Estimated respiratory rate,
          height=7cm,
          width=\textwidth,
      ]
      \addplot[
          only marks,
          mark=o,
          color=blue,
      ] table [col sep=comma, x=rr_est, y=rr_true] {data/results_vv_medium_scatter.csv};
      \addplot [gray,line width=0.2, domain=4:44, samples=2] {x};
      \end{axis}
    \end{tikzpicture}
  \end{subfigure}
  \caption{VitalLens estimated vitals vs. gold-standard true vitals on VV-Medium. Besides a few outliers, estimations closely match the true vitals.}
  \label{fig:two_scatterplots}
\end{figure}
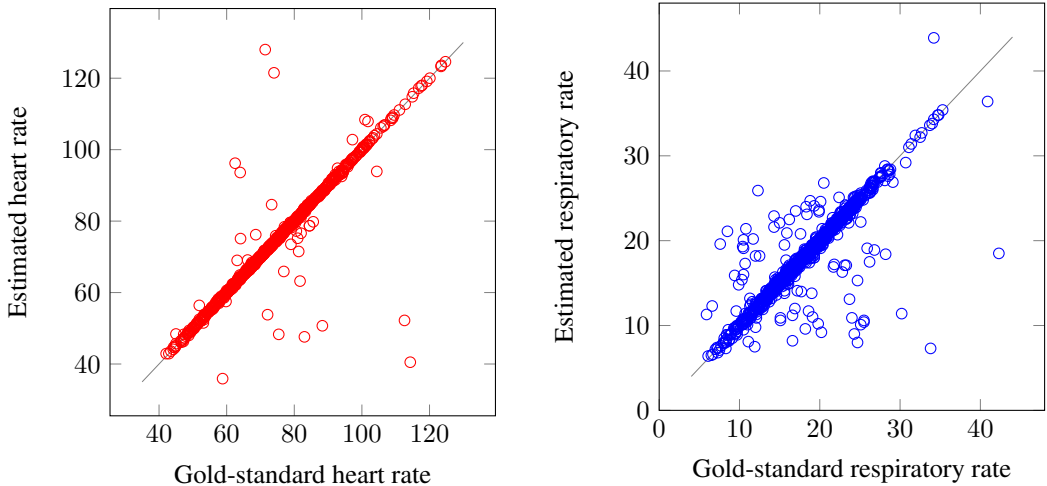

VitalLens beats all other listed methods for all considered evaluation metrics, while maintaining a competitive inference speed that is sufficient to run in real time at 30 fps.\footnote{The inference times are for a single video frame, excluding face detection but including all other computation, such as resizing. We measure these times using our own implementations as an average across 1,000 iterations on a workstation with Nvidia GeForce RTX 3080 and AMD Ryzen 7 3700X.}
The next best predictions for heart rate and respiratory rate also came from learning-based models MTTS-CAN and DeepPhys, respectively.
Note that this dataset has a large variety of participants and their demographics, but low amounts of participant and camera movement.

\begin{table}[h!]
 	\caption{Vitals estimation results on PROSIT test set}
 	\label{tab:results-prosit-test}
 	\centering
  	\begin{filecontents*}{data/results_prosit_test.csv}
method,hr_mae,pulse_snr,pulse_cor,rr_mae,resp_snr,resp_cor
G,27.70621927234887,-15.007110314059164,0.223712126011746,0.0,0.0,0.0
CHROM,24.06345975579728,-11.66658396952486,0.16333614514679298,0.0,0.0,0.0
POS,17.794871428491465,-11.943671900663842,0.27572711757170243,0.0,0.0,0.0
DeepPhys,29.52925589667416,-12.40852812467118,0.11496435026694374,4.950878731714404,-4.090759283881323,0.39775480749477865
MTTS-CAN,13.904032520228883,-6.726609133808825,0.33284411119368246,5.952886274980931,-6.478681530899787,0.23265762258675227
VitalLens,4.622805987010308,2.8313874453381995,0.6711940631318561,3.764053007131582,-1.2328959859592052,0.46940915572272246
\end{filecontents*}
\csvreader[
		tabular=lcccccc,
    	table head=\toprule Method & \multicolumn{1}{c}{Heart rate} & \multicolumn{2}{c}{Pulse wave} & \multicolumn{1}{c}{Respiratory rate} & \multicolumn{2}{c}{Respiration wave} \\ & MAE $\downarrow$ & SNR $\uparrow$ & $r \uparrow$ & MAE $\downarrow$ & SNR $\uparrow$ & $r \uparrow$ \\ \midrule,
    	table foot=\bottomrule,
    	late after line=\\,
    	before reading={\catcode`\%=12}, after reading={\catcode`\%=14}
  	]{data/results_prosit_test.csv}{}%
  {\csvcoli & \printMetric{\csvcolii} & \printMetric{\csvcoliii} & \printMetric{\csvcoliv} & \printMetric{\csvcolv} & \printMetric{\csvcolvi} & \printMetric{\csvcolvii} }
\end{table}

The PROSIT test set has a lower variety in participants and their demographics, but it includes both participant and camera movement.
Results for the PROSIT test set are reported in Table \ref{tab:results-prosit-test} -- again, VitalLens beats all listed methods for all considered evaluation metrics.
However, we observe that here, the absolute errors achieved by VitalLens for heart rate and respiratory rate are 4.62 bpm and 3.76 bpm, respectively -- markedly higher than for VV-Medium.
This is a first hint that factors like participant and camera movement are the main drivers of lower estimation accuracies. 

\begin{table}[h!]
 	\caption{Vitals estimation results on VV-Africa-Small test set}
 	\label{tab:results-vv-africa-small-test}
 	\centering
 	\begin{threeparttable}
  	\begin{filecontents*}{data/results_vv_africa_test.csv}
method,hr_mae,pulse_snr,pulse_cor,rr_mae,resp_snr,resp_cor
G,24.504919118798853,-12.480727370327347,0.1313409937161293,0.0,0.0,0.0
CHROM,18.900660412875126,-8.026297876060095,0.22567418238307554,0.0,0.0,0.0
POS,10.630417127774592,-6.203229008185227,0.1506948026221236,0.0,0.0,0.0
DeepPhys,26.787079682863304,-8.537079603947523,0.2890606731205406,2.572077474625463,3.5590268642424774,0.7507425392562387
MTTS-CAN,5.794236259700629,-2.906035543450423,0.5140226500265757,4.096137699265992,-2.7224252141408534,0.33718975800349194
VitalLens,1.7080614467346558,9.052028006606443,0.7879216534797234,2.426073502296371,4.4023894161816575,0.8135646744867556
\end{filecontents*}
\csvreader[
		tabular=lcccccc,
    	table head=\toprule Method & \multicolumn{1}{c}{Heart rate} & \multicolumn{2}{c}{Pulse wave} & \multicolumn{1}{c}{Respiratory rate\tnote{a}} & \multicolumn{2}{c}{Respiration wave\tnote{a}} \\ & MAE $\downarrow$ & SNR $\uparrow$ & $r \uparrow$ & MAE $\downarrow$ & SNR $\uparrow$ & $r \uparrow$ \\ \midrule,
    	table foot=\bottomrule,
    	late after line=\\,
    	before reading={\catcode`\%=12}, after reading={\catcode`\%=14}
  	]{data/results_vv_africa_test.csv}{}%
  {\csvcoli & \printMetric{\csvcolii} & \printMetric{\csvcoliii} & \printMetric{\csvcoliv} & \printMetric{\csvcolv} & \printMetric{\csvcolvi} & \printMetric{\csvcolvii} }
  \begin{tablenotes}
	\item[a] Missing respiratory signals were synthetically created and manually verified.
	\end{tablenotes}
  \end{threeparttable}
\end{table}

Finally, Table \ref{tab:results-vv-africa-small-test} lists the results for the VV-Africa-Small test set.
The results paint a similar picture as the two previous datasets, with VitalLens beating out the other methods.
In comparison to VV-Medium, we observe that the errors are generally higher - in this case, this may be caused by the demographic differences between the participants and more challenging lighting conditions.
We further investigate this in the following sections.

\subsection{Which factors impact estimation performance?}

We consider a number of factors that may impact estimation performance of pulse and respiration.

\paragraph{Demographic factors:}

\begin{enumerate}[align=left]
	\item[\texttt{age}:] The age of the participant.
	\item[\texttt{gender\_male}:] Dummy variable indicating whether the participant is male.
	\item[\texttt{hr} / \texttt{rr}:] The heart rate and respiratory rate of the participant.
	\item[\texttt{skin\_type\_x}:] Dummy variable indicating whether the participant has skin type x. Base case is skin type 1.
\end{enumerate}

\paragraph{Behavioral/environmental factors:}

\begin{enumerate}[align=left]
	\item[\texttt{camera\_stationary}:] Indicates whether the camera is stationary.
	\item[\texttt{illuminance\_var}:] Measures how much the illuminance of the participant's faces varies throughout a chunk, in interval [0,1]. Derived from the luma component of the pixels within the participant's face. Partially caused by movement ($r = 0.2$ for PROSIT).
	\item[\texttt{movement}:] Measures how much the participant moved throughout the chunk, in interval [0,1]. Derived from the movement of facial landmarks.
\end{enumerate}

\subsubsection{Pulse wave}

To investigate which factors have a significant impact, we conduct OLS regressions using them to predict the SNR for pulse wave estimation.
We perform two separate regressions, one using the chunks of the VV-Medium dataset and one for the PROSIT test set - the results are reported in Tables \ref{tab:vv-hr-regression-vv} and \ref{tab:vv-hr-regression-prosit}.

\begin{table}[h!]
\begin{center}
\caption{Regression analysis of factors affecting pulse wave estimation on VV-Medium}
\label{tab:vv-hr-regression-vv}
\begin{tabular}{lclc}
\toprule
\textbf{Dep. Variable:}          &  Pulse wave SNR  & \textbf{  R-squared:         } &     0.264   \\
\textbf{Model:}                  &       OLS        & \textbf{  Adj. R-squared:    } &     0.257   \\
\textbf{Method:}                 &  Least Squares   & \textbf{  F-statistic:       } &     39.27   \\
\textbf{No. Observations:}       &        1108      & \textbf{  Prob (F-statistic):} &  2.40e-66   \\
\textbf{Df Residuals:}           &        1097      & \textbf{  AIC:               } &     7118.   \\
\textbf{Df Model:}               &          10      & \textbf{  BIC:               } &     7174.   \\
\bottomrule
\end{tabular}
\begin{tabular}{lcccccc}
                                 & \textbf{coef} & \textbf{std err} & \textbf{t} & \textbf{P$> |$t$|$} & \textbf{[0.025} & \textbf{0.975]}  \\
\midrule
\textbf{intercept}               &       9.7449  &        1.575     &     6.189  &         0.000        &        6.655    &       12.834     \\
\textbf{age}                     &       0.1135  &        0.010     &    11.919  &         0.000        &        0.095    &        0.132     \\
\textbf{illuminance\_var}        &     -22.3650  &        4.096     &    -5.460  &         0.000        &      -30.403    &      -14.327     \\
\textbf{movement}                &      -5.3373  &        3.179     &    -1.679  &         0.093        &      -11.576    &        0.901     \\
\textbf{hr}                      &      -0.0283  &        0.014     &    -2.046  &         0.041        &       -0.055    &       -0.001     \\
\textbf{gender\_male}            &       0.2530  &        0.377     &     0.672  &         0.502        &       -0.486    &        0.992     \\
\textbf{skin\_type\_2}  			 &      -0.1885  &        1.024     &    -0.184  &         0.854        &       -2.199    &        1.822     \\
\textbf{skin\_type\_3}  			 &       0.4203  &        1.104     &     0.381  &         0.703        &       -1.745    &        2.586     \\
\textbf{skin\_type\_4}  			 &       1.4931  &        1.161     &     1.286  &         0.199        &       -0.785    &        3.772     \\
\textbf{skin\_type\_5}  			 &      -2.6020  &        1.140     &    -2.282  &         0.023        &       -4.839    &       -0.365     \\
\textbf{skin\_type\_6}  			 &      -7.9805  &        1.241     &    -6.429  &         0.000        &      -10.416    &       -5.545     \\
\bottomrule
\end{tabular}
\end{center}
\end{table}

For our regression analysis on the VV-Medium, we use all available factors.
The regression equation is significant and explains 26.4\% of the variance of the pulse wave estimation SNR.
\texttt{age}, \texttt{hr}, \texttt{illuminance\_var}, and \texttt{skin\_type\_x} are shown to have significant effects at the 5\% level:

\begin{itemize}
	\item \texttt{age} has a weak positive effect, meaning that estimations were slightly more accurate for older participants,
	\item \texttt{hr} has a weak negative effect, meaning that estimations were slightly less accurate for participants with higher heart rates,
	\item \texttt{illuminance\_var} had a strong negative effect, meaning that higher variance in facial illuminance of participants led to less accurate estimations, and
	\item \texttt{skin\_type\_5} and especially \texttt{skin\_type\_6} had negative effects, meaning that pulse wave estimation for participants with skin types 5 and especially 6 was less accurate.
\end{itemize}

It is worth noting that \texttt{movement} did not have a significant effect, potentially because participants generally do not move much for this dataset.

\begin{table}[h!]
\begin{center}
\caption{Regression analysis of factors affecting pulse wave estimation on PROSIT test set}
\label{tab:vv-hr-regression-prosit}
\begin{tabular}{lclc}
\toprule
\textbf{Dep. Variable:}          &  Pulse wave SNR  & \textbf{  R-squared:         } &     0.103   \\
\textbf{Model:}                  &       OLS        & \textbf{  Adj. R-squared:    } &     0.102   \\
\textbf{Method:}                 &  Least Squares   & \textbf{  F-statistic:       } &     53.82   \\
\textbf{No. Observations:}       &        1403      & \textbf{  Prob (F-statistic):} &  6.40e-33   \\
\textbf{Df Residuals:}           &        1399      & \textbf{  AIC:               } & 1.040e+04   \\
\textbf{Df Model:}               &           3      & \textbf{  BIC:               } & 1.042e+04   \\
\bottomrule
\end{tabular}
\begin{tabular}{lcccccc}
                                 & \textbf{coef} & \textbf{std err} & \textbf{t} & \textbf{P$> |$t$|$} & \textbf{[0.025} & \textbf{0.975]}  \\
\midrule
\textbf{intercept}               &       7.2056  &        0.669     &    10.775  &         0.000        &        5.894    &        8.517     \\
\textbf{illuminance\_var} 		 &     -26.4655  &        3.216     &    -8.230  &         0.000        &      -32.774    &      -20.157     \\
\textbf{movement}       			 &     -14.3836  &        1.687     &    -8.525  &         0.000        &      -17.693    &      -11.074     \\
\textbf{camera\_stationary}   	 &       0.1855  &        0.597     &     0.311  &         0.756        &       -0.985    &        1.356     \\
\bottomrule
\end{tabular}
\end{center}
\end{table}

Since there are not many unique participants in the PROSIT test set, we only consider the behavioral and environmental factors for the next regression.
Looking at the results, the regression equation itself as well as \texttt{illuminance\_var} and \texttt{movement} have significant effects at the 5\% level: 

\begin{itemize}
	\item \texttt{illuminance\_var} has a strong negative effect, meaning that estimations became less accurate with higher variance in facial illuminance of participants.
	\item \texttt{movement} has a negative effect, meaning that estimations were less accurate when participants moved more.
\end{itemize}

These behavioral factors are collectively able to explain 10.3\% of the variance of the pulse wave estimation SNR.
It is worth noting that \texttt{camera\_stationary} did not have a significant effect, which indicates that our rPPG model has some degree of robustness against camera movement.

Overall, the results confirm that the behavioral factors of participant movement and variation in the illuminance of the participant's face dominate the results with a significant negative impact on estimation performance.
It is interesting to note that at least for these datasets, the effect of illuminance variation seems to have a greater impact than participant movement.
Our model performance appears to be slightly impacted by the heart rate itself.
We also confirm previous findings that rPPG models can struggle with darker skin types, but cannot confirm findings that estimation is more accurate for females \cite{nowara2020meta}.
Additionally, we observe that estimation appears to be more accurate for older participants.

\subsubsection{Respiration wave}

For our analysis of the impact on respiration wave estimation, we conduct another two OLS regressions.
The results are reported in Table \ref{tab:vv-rr-regression-vv} and \ref{tab:vv-rr-regression-prosit}.

\begin{table}[h!]
\begin{center}
\caption{Regression analysis of factors affecting respiratory rate estimation on VV-Medium}
\label{tab:vv-rr-regression-vv}
\begin{tabular}{lclc}
\toprule
\textbf{Dep. Variable:}          &  Resp wave SNR   & \textbf{  R-squared:         } &     0.071   \\
\textbf{Model:}                  &       OLS        & \textbf{  Adj. R-squared:    } &     0.062   \\
\textbf{Method:}                 &  Least Squares   & \textbf{  F-statistic:       } &     8.340   \\
\textbf{No. Observations:}       &        1108      & \textbf{  Prob (F-statistic):} &  3.71e-13   \\
\textbf{Df Residuals:}           &        1097      & \textbf{  AIC:               } &     7352.   \\
\textbf{Df Model:}               &          10      & \textbf{  BIC:               } &     7407.   \\
\bottomrule
\end{tabular}
\begin{tabular}{lcccccc}
                                 & \textbf{coef} & \textbf{std err} & \textbf{t} & \textbf{P$> |$t$|$} & \textbf{[0.025} & \textbf{0.975]}  \\
\midrule
\textbf{intercept}               &       8.6117  &        1.349     &     6.385  &         0.000        &        5.965    &       11.258     \\
\textbf{age}            			 &       0.0674  &        0.010     &     6.559  &         0.000        &        0.047    &        0.088     \\
\textbf{illuminance\_var} 		 &      -4.6049  &        4.564     &    -1.009  &         0.313        &      -13.560    &        4.350     \\
\textbf{movement}       			 &     -13.8675  &        3.536     &    -3.922  &         0.000        &      -20.805    &       -6.930     \\
\textbf{rr}          			 &      -0.1615  &        0.040     &    -4.056  &         0.000        &       -0.240    &       -0.083     \\
\textbf{gender\_male}   			 &       0.7883  &        0.407     &     1.939  &         0.053        &       -0.009    &        1.586     \\
\textbf{skin\_type\_2}  			 &      -0.8223  &        1.139     &    -0.722  &         0.470        &       -3.057    &        1.413     \\
\textbf{skin\_type\_3}  			 &      -0.5369  &        1.230     &    -0.437  &         0.663        &       -2.950    &        1.876     \\
\textbf{skin\_type\_4}  			 &       0.1492  &        1.296     &     0.115  &         0.908        &       -2.395    &        2.693     \\
\textbf{skin\_type\_5}  			 &       1.1431  &        1.269     &     0.901  &         0.368        &       -1.346    &        3.632     \\
\textbf{skin\_type\_6}  			 &      -0.3998  &        1.383     &    -0.289  &         0.773        &       -3.114    &        2.314     \\
\bottomrule
\end{tabular}
\end{center}
\end{table}

For our regression analysis on VV-Medium, we use all available factors.
The regression equation is significant but explains only 7.1\% of the variance of the respiratory wave estimation SNR.
\texttt{age}, \texttt{movement}, and \texttt{rr} are shown to have significant effects at the 5\% level:

\begin{itemize}
	\item \texttt{age} has a weak positive effect, meaning that estimations were slightly more accurate for older participants,
	\item \texttt{movement} had a strong negative effect, meaning that higher participant movement led to less accurate estimations, and
	\item \texttt{rr} has a weak negative effect, meaning that estimations were slightly less accurate for participants with higher respiratory rates.
\end{itemize}

\begin{table}[h!]
\begin{center}
\caption{Regression analysis of factors affecting respiratory rate estimation on PROSIT test set}
\label{tab:vv-rr-regression-prosit}
\begin{tabular}{lclc}
\toprule
\textbf{Dep. Variable:}          &  Resp wave SNR   & \textbf{  R-squared:         } &     0.068   \\
\textbf{Model:}                  &       OLS        & \textbf{  Adj. R-squared:    } &     0.066   \\
\textbf{Method:}                 &  Least Squares   & \textbf{  F-statistic:       } &     33.96   \\
\textbf{No. Observations:}       &        1403      & \textbf{  Prob (F-statistic):} &  3.46e-21   \\
\textbf{Df Residuals:}           &        1399      & \textbf{  AIC:               } &     9735.   \\
\textbf{Df Model:}               &           3      & \textbf{  BIC:               } &     9756.   \\
\bottomrule
\end{tabular}
\begin{tabular}{lcccccc}
                                 & \textbf{coef} & \textbf{std err} & \textbf{t} & \textbf{P$> |$t$|$} & \textbf{[0.025} & \textbf{0.975]}  \\
\midrule
\textbf{intercept}               &       2.4654  &        0.527     &     4.676  &         0.000        &        1.431    &        3.500     \\
\textbf{illuminance\_var}  		 &     -11.0138  &        2.535     &    -4.344  &         0.000        &      -15.987    &       -6.040     \\
\textbf{movement}                &     -11.9294  &        1.330     &    -8.968  &         0.000        &      -14.539    &       -9.320     \\
\textbf{camera\_stationary}      &      -1.3251  &        0.471     &    -2.816  &         0.005        &       -2.248    &       -0.402     \\
\bottomrule
\end{tabular}
\end{center}
\end{table}

For the regression on the PROSIT test set, we again focus on the behavioral and environmental factors.
The regression itself and all considered factors have significant effects at the 5\% level:

\begin{itemize}
	\item \texttt{camera\_stationary} has a weak negative effect, meaning that estimations were slightly less accurate when camera was stationary,
	\item \texttt{illuminance\_var} has a strong negative effect, meaning that estimations were less accurate when participant illuminance varied more, and
	\item \texttt{movement} had a strong negative effect, meaning that higher participant movement led to less accurate estimations.
\end{itemize}

The factors are collectively able to explain 6.8\% of the variance of the respiratory wave estimation SNR.
As expected, participant movement degrades the estimation quality for the respiratory rate.
It is surprising that variation in the illuminance of the participant's face impacts the estimation to a similar degree.
Also, the finding that estimations were more accurate when the camera was not stationary is counterintuitive.
One possible explanation is that VitalLens learns to understand that for some participants, repetitive movement of the (handheld) camera actually reveals information about respiration.

Based on the results for both pulse and respiration, we advise users of VitalLens that for best performance they should (i) hold still, and (ii) make sure the face is evenly illuminated.

\subsubsection{Impact of participant movement}

\pgfplotstableread[col sep=comma]{data/impact_movement.csv}\datatable

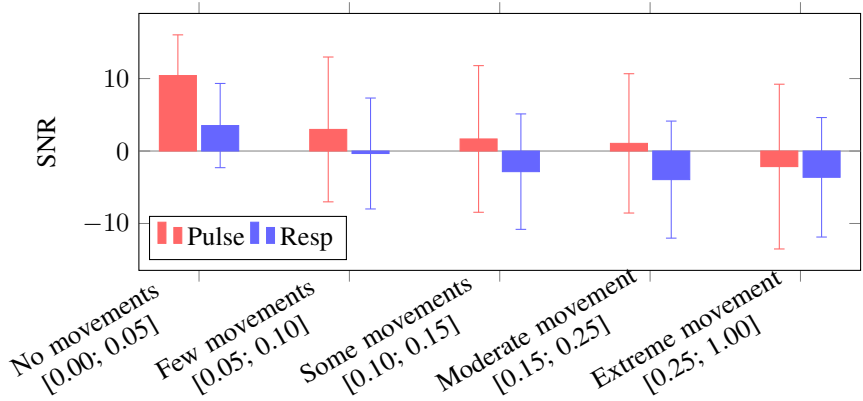
\begin{figure}[h!]
  \centering
  \begin{tikzpicture}
    \begin{axis}[
        ylabel=SNR,
        height=5cm,
        width=.8\textwidth,
        legend style={
            at={(0.014,.04)},
            anchor=south west,
            legend columns=-1
        },
        ybar,
        bar width=14.,
        xtick=data,
        xticklabels from table={\datatable}{bin},
        xticklabel style={rotate=30, anchor=east, align=center, text width=3cm},
    ]
    \draw[gray, line width=0.2] ({rel axis cs:0,0}|-{axis cs:0,0}) -- ({rel axis cs:1,0}|-{axis cs:0,0});
    \addplot+[
    	red!60,
        ybar,
        error bars/.cd,
        y dir=both,
        y explicit,
    ] table[x expr=\coordindex, y=pulse_snr_mean, y error=pulse_snr_sd] {\datatable};
    \addlegendentry{Pulse};
    \addplot+[
        blue!60,
        ybar,
        error bars/.cd,
        y dir=both,
        y explicit,
    ] table[x expr=\coordindex, y=resp_snr_mean, y error=resp_snr_sd] {\datatable};
    \addlegendentry{Resp};
    \end{axis}
  \end{tikzpicture}
  \caption{Signal-to-noise ratio for pulse and respiration on the PROSIT test set, grouped by different levels of participant movement.}
  \label{fig:impact-movement}
\end{figure}

According to the regression analysis, increasing amounts of participant movement lead to worse estimation performances for both heart rate and respiratory rate.
We explore this effect in Figure \ref{fig:impact-movement} by looking the average SNR for different buckets of participant movement on the PROSIT test set.
It is evident that SNR values indeed decrease with increased movement.
There appears to be a large drop-off especially going from chunks that have no movement to those that have a few movements.

\subsubsection{Impact of participant illuminance variation}

\pgfplotstableread[col sep=comma]{data/impact_illuminance.csv}\datatable

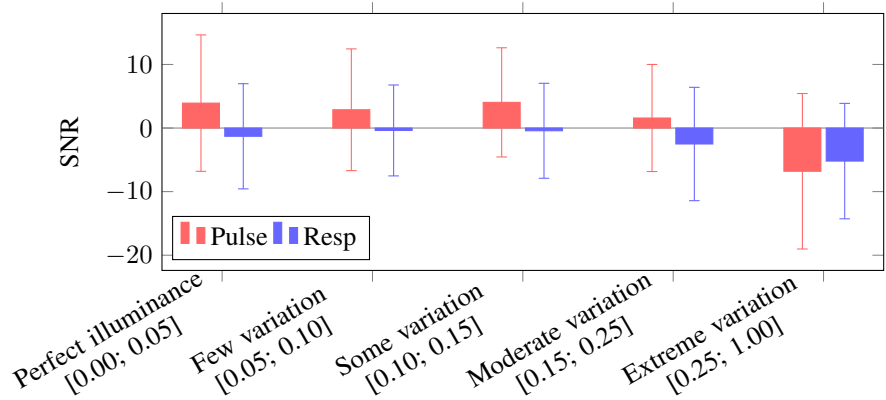
\begin{figure}[h!]
  \centering
  \begin{tikzpicture}
    \begin{axis}[
        ylabel=SNR,
        height=5cm,
        width=.8\textwidth,
        legend style={
            at={(0.014,.04)},
            anchor=south west,
            legend columns=-1
        },
        ybar,
        bar width=14.,
        xtick=data,
        xticklabels from table={\datatable}{bin},
        xticklabel style={rotate=30, anchor=east, align=center, text width=3cm},
    ]
    \draw[gray, line width=0.2] ({rel axis cs:0,0}|-{axis cs:0,0}) -- ({rel axis cs:1,0}|-{axis cs:0,0});
    \addplot+[
    	red!60,
        ybar,
        error bars/.cd,
        y dir=both,
        y explicit,
    ] table[x expr=\coordindex, y=pulse_snr_mean, y error=pulse_snr_sd] {\datatable};
    \addlegendentry{Pulse};
    \addplot+[
        blue!60,
        ybar,
        error bars/.cd,
        y dir=both,
        y explicit,
    ] table[x expr=\coordindex, y=resp_snr_mean, y error=resp_snr_sd] {\datatable};
    \addlegendentry{Resp};
    \end{axis}
  \end{tikzpicture}
  \caption{Signal-to-noise ratio for pulse and respiration on the PROSIT test set, grouped by different levels of participant illuminance variation.}
  \label{fig:impact-illuminance}
\end{figure}

The regression analysis showed that increasing amounts of variation in participant illuminance also led to worse estimation performances for both heart rate and respiratory rate.
We explore this effect in Figure \ref{fig:impact-illuminance} by looking at the average SNR for different buckets of variation in participant illuminance on the PROSIT test set.
This figure reveals that especially when going from chunks with moderate illuminance variation to those with extreme illuminance variation, we see a large drop-off in estimation accuracy.

\subsubsection{Impact of participant age}

\pgfplotstableread[col sep=comma]{data/impact_age.csv}\datatable

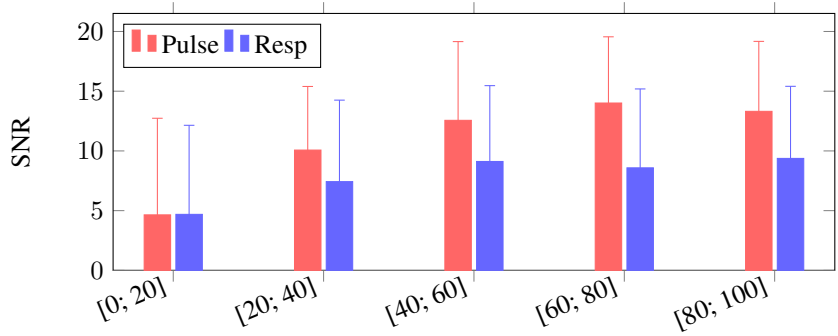
\begin{figure}[h!]
  \centering
  \begin{tikzpicture}
    \begin{axis}[
        x tick label style={
            /pgf/number format/1000 sep=},
        ylabel=SNR,
        height=5cm,
        width=.8\textwidth,
        legend style={
            at={(0.014,.96)},
            anchor=north west,
            legend columns=-1
        },
        ymin=0,
        ybar,
        bar width=10.,
        xtick=data,
        xticklabels from table={\datatable}{bin},
        xticklabel style={rotate=22.5, anchor=east},
    ]
    \draw[gray, line width=0.2] ({rel axis cs:0,0}|-{axis cs:0,0}) -- ({rel axis cs:1,0}|-{axis cs:0,0});
    \addplot+[
    	red!60,
        ybar,
        error bars/.cd,
        y dir=both,
        y explicit,
    ] table[x expr=\coordindex, y=pulse_snr_mean, y error=pulse_snr_sd] {\datatable};
    \addlegendentry{Pulse};
    \addplot+[
        blue!60,
        ybar,
        error bars/.cd,
        y dir=both,
        y explicit,
    ] table[x expr=\coordindex, y=resp_snr_mean, y error=resp_snr_sd] {\datatable};
    \addlegendentry{Resp};
    \end{axis}
  \end{tikzpicture}
  \caption{Signal-to-noise ratio for pulse and respiration on VV-Medium, grouped by different participant ages.}
  \label{fig:impact-age}
\end{figure}

Since the regression results indicated that VitalLens is slightly more accurate overall for older participants, we explore this effect in more detail as well.
Figure \ref{fig:impact-age} indeed shows that especially for participants under 20 years of age, the SNR for both pulse and respiration is tends to be lower.

\subsubsection{Impact of participant skin type}

\pgfplotstableread[col sep=comma]{data/impact_skin_type.csv}\datatable

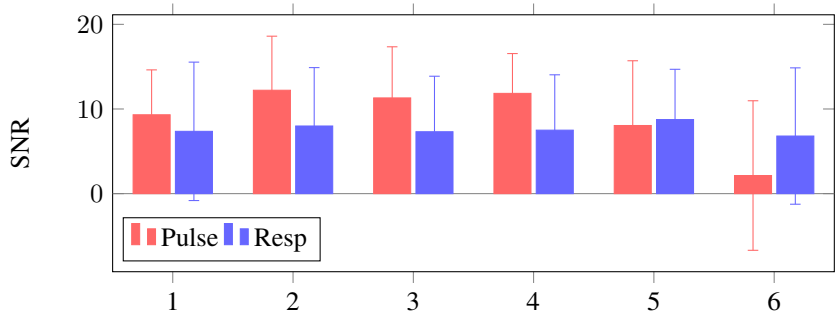
\begin{figure}[h!]
  \centering
  \begin{tikzpicture}
    \begin{axis}[
        ylabel=SNR,
        height=5cm,
        width=.8\textwidth,
        legend style={
            at={(0.014,.04)},
            anchor=south west,
            legend columns=-1
        },
        ybar,
        bar width=14.,
        xtick=data,
        xticklabels from table={\datatable}{bin},
        xticklabel style={align=center, text width=3cm},
    ]
    \draw[gray, line width=0.2] ({rel axis cs:0,0}|-{axis cs:0,0}) -- ({rel axis cs:1,0}|-{axis cs:0,0});
    \addplot+[
    	red!60,
        ybar,
        error bars/.cd,
        y dir=both,
        y explicit,
    ] table[x expr=\coordindex, y=pulse_snr_mean, y error=pulse_snr_sd] {\datatable};
    \addlegendentry{Pulse};
    \addplot+[
        blue!60,
        ybar,
        error bars/.cd,
        y dir=both,
        y explicit,
    ] table[x expr=\coordindex, y=resp_snr_mean, y error=resp_snr_sd] {\datatable};
    \addlegendentry{Resp};
    \end{axis}
  \end{tikzpicture}
  \caption{Signal-to-noise ratio for pulse and respiration on VV-Medium, grouped by different participant skin types.}
  \label{fig:impact-skin-type}
\end{figure}

According to the regression analysis on VV-Medium, the pulse wave estimation had a lower accuracy especially for participants with skin type 6.
Here, we seek to further explore this effect.
We start by grouping the chunks in VV-Medium by skin type and then plot the average estimation SNR for both pulse wave and respiration wave estimation in Figure \ref{fig:impact-skin-type}.
Indeed, we observe that consistent with our previous regression analysis, there is a drop-off in the SNR between skin type 5 and 6 for pulse, while there is no obvious effect visible for respiration.

\pgfplotstableread[col sep=comma]{data/impact_skin_type_comparison.csv}\datatable

\begin{figure}[h!]
  \centering
  \begin{tikzpicture}
    \begin{axis}[
        ylabel=SNR,
        height=5cm,
        width=.8\textwidth,
        legend style={
            at={(0.014,.04)},
            anchor=south west,
            legend columns=-1
        },
        ybar,
        bar width=14.,
        xtick=data,
        xticklabels from table={\datatable}{bin},
        xticklabel style={align=center, text width=3cm},
    ]
    \draw[gray, line width=0.2] ({rel axis cs:0,0}|-{axis cs:0,0}) -- ({rel axis cs:1,0}|-{axis cs:0,0});
    \addplot+[
    	red!60,
        ybar,
        error bars/.cd,
        y dir=both,
        y explicit,
    ] table[x expr=\coordindex, y=pulse_snr_mean_vl, y error=pulse_snr_sd_vl] {\datatable};
    \addlegendentry{VitalLens};
    \addplot+[
        green!60,
        ybar,
        error bars/.cd,
        y dir=both,
        y explicit,
    ] table[x expr=\coordindex, y=pulse_snr_mean_vl_64, y error=pulse_snr_sd_vl_64] {\datatable};
    \addlegendentry{VitalLens*};
    \end{axis}
  \end{tikzpicture}
  \caption{Comparing signal-to-noise ratio between VitalLens and VitalLens* on VV-Medium, grouped by participant skin types. VitalLens* was solely trained on PROSIT, while VitalLens was trained on both PROSIT and VV-Africa-Small.}
  \label{fig:impact-skin-type-comparison}
\end{figure}
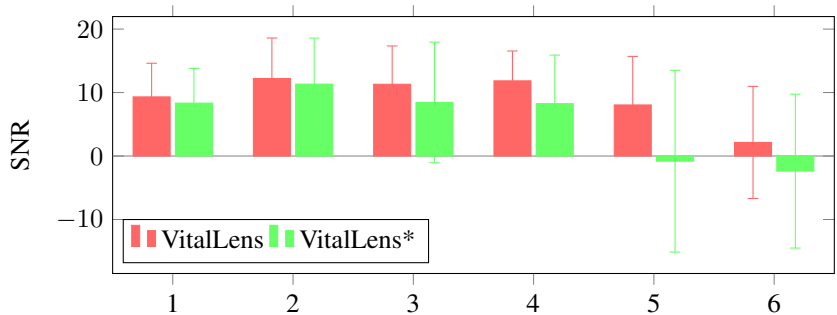

Next, we want to investigate to what degree the inclusion of VV-Africa-Small in our training data lessened this skin type effect.
For this purpose, we compare VitalLens with \textit{VitalLens*}, a version of VitalLens that is solely trained on PROSIT.
The training set of PROSIT by itself only includes 2 individuals of skin type 6, a number that is massively increased in the combined training data.
Both VitalLens and VitalLens* are then evaluated on VV-Medium.
Figure \ref{fig:impact-skin-type-comparison} compares the results, again grouped by skin type.
It is evident that the addition of VV-Africa-Small led to a sizeable improvement mainly for skin types 5 and 6, but also skin types 3 and 4.

\begin{figure}[h!]
  \centering
  \begin{tikzpicture}
    \begin{axis}[
        ylabel=SNR,
        height=5cm,
        width=.8\textwidth,
        legend style={
            at={(0.014,.04)},
            anchor=south west,
            legend columns=-1
        },
        ybar,
        bar width=14.,
        xtick=data,
        xticklabels from table={\datatable}{bin},
        xticklabel style={align=center, text width=3cm},
    ]
    \draw[gray, line width=0.2] ({rel axis cs:0,0}|-{axis cs:0,0}) -- ({rel axis cs:1,0}|-{axis cs:0,0});
    \addplot+[
    	red!60,
        ybar,
        error bars/.cd,
        y dir=both,
        y explicit,
    ] table[x expr=\coordindex, y=pulse_snr_mean_vl, y error=pulse_snr_sd_vl] {\datatable};
    \addlegendentry{VitalLens};
    \addplot+[
        green!60,
        ybar,
        error bars/.cd,
        y dir=both,
        y explicit,
    ] table[x expr=\coordindex, y=pulse_snr_mean_pos, y error=pulse_snr_sd_pos] {\datatable};
    \addlegendentry{POS};
    \end{axis}
  \end{tikzpicture}
  \caption{Comparing signal-to-noise ratio between VitalLens and POS on VV-Medium, grouped by participant skin types.}
  \label{fig:impact-skin-type-comparison-pos}
\end{figure}
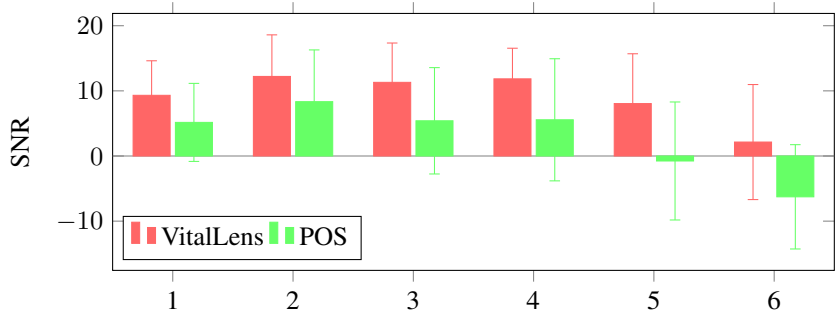

Finally, we compare the impact of skin types on pulse wave estimation between our learned model VitalLens and the best-performing handcrafted model POS.
As visible in Figure \ref{fig:impact-skin-type-comparison-pos}, the impact of skin types in POS is even worse than on VitalLens*, with an even larger drop-off in SNR between skin types 5 and 6.

We can conclude from this analysis that issues arising from the lower pulse signal strength in rPPG for darker skin types can at least partially be overcome by learning-based rPPG models which use training data that actually includes a significant amount of individuals with those skin types.

\subsubsection{Impact of participant vitals}

Finally, we investigate the impact of the vitals themselves on estimation performance.
Although we took steps to prevent this, the distribution of vitals in the training set may have biased VitalLens towards predicting the average values found in the training set.
The regressions found that estimations generally became slightly less accurate for individuals with higher heart rates and respiratory rates.
Figures \ref{fig:impact-hr} and \ref{fig:impact-rr} show the SNR for pulse and respiration estimation, grouped by different heart rate and respiratory rates respectively.

\pgfplotstableread[col sep=comma]{data/impact_hr.csv}\datatable

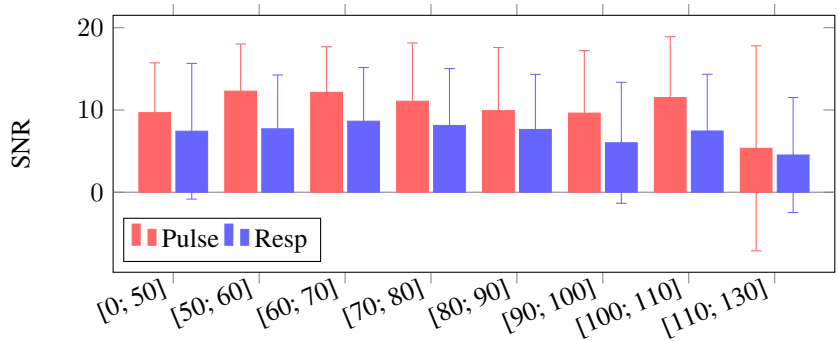
\begin{figure}[h!]
  \centering
  \begin{tikzpicture}
    \begin{axis}[
        x tick label style={
            /pgf/number format/1000 sep=},
        ylabel=SNR,
        height=5cm,
        width=.8\textwidth,
        legend style={
            at={(0.014,.04)},
            anchor=south west,
            legend columns=-1
        },
        ybar,
        bar width=12.,
        xtick=data,
        xticklabels from table={\datatable}{bin},
        xticklabel style={rotate=22.5, anchor=east},
    ]
    \draw[gray, line width=0.2] ({rel axis cs:0,0}|-{axis cs:0,0}) -- ({rel axis cs:1,0}|-{axis cs:0,0});
    \addplot+[
    	red!60,
        ybar,
        error bars/.cd,
        y dir=both,
        y explicit,
    ] table[x expr=\coordindex, y=pulse_snr_mean, y error=pulse_snr_sd] {\datatable};
    \addlegendentry{Pulse};
    \addplot+[
        blue!60,
        ybar,
        error bars/.cd,
        y dir=both,
        y explicit,
    ] table[x expr=\coordindex, y=resp_snr_mean, y error=resp_snr_sd] {\datatable};
    \addlegendentry{Resp};
    \end{axis}
  \end{tikzpicture}
  \caption{Signal-to-noise ratio for pulse and respiration on VV-Medium, grouped by different participant heart rate values.}
  \label{fig:impact-hr}
\end{figure}

\pgfplotstableread[col sep=comma]{data/impact_rr.csv}\datatable

\begin{figure}[h!]
  \centering
  \begin{tikzpicture}
    \begin{axis}[
        x tick label style={
            /pgf/number format/1000 sep=},
        ylabel=SNR,
        height=5cm,
        width=.8\textwidth,
        legend style={
            at={(0.014,.04)},
            anchor=south west,
            legend columns=-1
        },
        ybar,
        bar width=14.,
        ymin=-7,
        xtick=data,
        xticklabels from table={\datatable}{bin},
        xticklabel style={rotate=22.5, anchor=east},
    ]
    \draw[gray, line width=0.2] ({rel axis cs:0,0}|-{axis cs:0,0}) -- ({rel axis cs:1,0}|-{axis cs:0,0});
    \addplot+[
    	red!60,
        ybar,
        error bars/.cd,
        y dir=both,
        y explicit,
    ] table[x expr=\coordindex, y=pulse_snr_mean, y error=pulse_snr_sd] {\datatable};
    \addlegendentry{Pulse};
    \addplot+[
        blue!60,
        ybar,
        error bars/.cd,
        y dir=both,
        y explicit,
    ] table[x expr=\coordindex, y=resp_snr_mean, y error=resp_snr_sd] {\datatable};
    \addlegendentry{Resp};
    \end{axis}
  \end{tikzpicture}
  \caption{Signal-to-noise ratio for pulse and respiration on VV-Medium, grouped by different participant respiratory rate values.}
  \label{fig:impact-rr}
\end{figure}
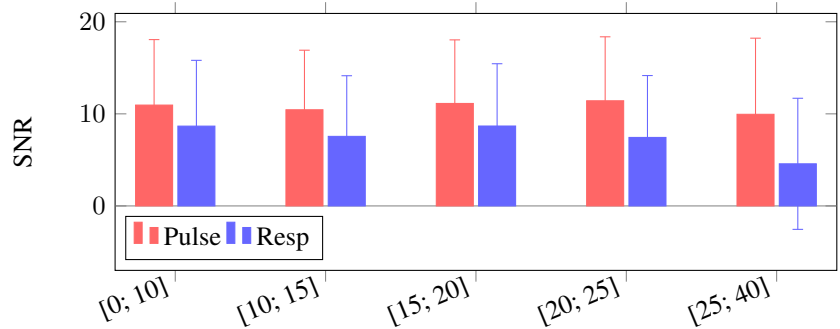

The figures show that generally, VitalLens does not appear to be biased by the distributions of heart rate and respiratory rate in the training data.
We do however note that pulse estimation became less accurate for the chunks with the highest heart rates, and respiration estimation became less accurate for the chunks with the highest respiration rates.
This consistent with the findings of the regression analysis.

\section{Conclusion}
\label{sec:conclusion}

VitalLens emerges as a novel solution in the domain of rPPG, showcasing its potential for accurate and real-time estimation of vital signs from selfie videos.
The extensive evaluation establishes its superiority over established methods, such as G, CHROM, POS, DeepPhys, and MTTS-CAN.
Notably, VitalLens outperforms these methods while maintaining fast inference times.
The thorough analysis of factors impacting performance reveals the significance of participant movement and illuminance variation, providing valuable insights for optimal usage.
Additionally, the inclusion of diverse training data to address skin type bias, contributes to the model's robustness.

\begin{ack}
We are thankful to Pieter-Jan Toye and Andrew Goodwin for the helpful suggestions.
This work was funded by Rouast Labs Pty Ltd.
\end{ack}

\bibliographystyle{plain}
\bibliography{paper}


\end{document}